\documentclass[runningheads]{llncs}

\usepackage{xcolor}
\usepackage{graphicx}
\usepackage{array}
\usepackage{tabularx}
\usepackage{makecell}
\usepackage{subcaption}

\begin{document}
\title{VAULT: A Mobile Mapping System for ROS 2-based Autonomous Robots} 
%
%
\author{Miguel Á. González-Santamarta\inst{1}\orcidID{0000-0002-7658-8600} \and
Francisco J. Rodrıguez-Lera\inst{1}\orcidID{0000-0002-8400-7079} \and
Vicente Matellan-Olivera\inst{1}\orcidID{0000-0001-7844-9658}}
\authorrunning{Miguel Á. González-Santamarta et al.}
%
\institute{Universidad de León, Av. Facultad de Veterinaria, 24007, León, Spain \\
\email{\{mgons,fjrodl,vmato\}@unileon.es}}
\maketitle              
\begin{abstract}
Localization plays a crucial role in the navigation capabilities of autonomous robots, and while indoor environments can rely on wheel odometry and 2D LiDAR-based mapping, outdoor settings such as agriculture and forestry, present unique challenges that necessitate real-time localization and consistent mapping. Addressing this need, this paper introduces the VAULT prototype, a ROS 2-based mobile mapping system (MMS) that combines various sensors to enable robust outdoor and indoor localization. The proposed solution harnesses the power of Global Navigation Satellite System (GNSS) data, visual-inertial odometry (VIO), inertial measurement unit (IMU) data, and the Extended Kalman Filter (EKF) to generate reliable 3D odometry. To further enhance the localization accuracy, Visual SLAM (VSLAM) is employed, resulting in the creation of a comprehensive 3D point cloud map. By leveraging these sensor technologies and advanced algorithms, the prototype offers a comprehensive solution for outdoor localization in autonomous mobile robots, enabling them to navigate and map their surroundings with confidence and precision. 


\keywords{Real-time Localization \and Mobile Mapping Systems \and Sensor Fusion \and 3D Mapping \and Visual SLAM \and Visual Odometry}
\end{abstract}
\section{Introduction}


Localization in the context of mobile mapping systems (MMS), refers to the process of enabling the creation of detailed and dynamic maps in which determining the precise position and orientation of the MMS platform (e.g., a device mounted on the robot system) in the real world. This localization data is especially important in autonomous mobile robots where robots use their current position to navigate through the environment.



Localization is one of the fundamental competencies an autonomous robot requires, as knowledge of the robot's location is essential for making decisions about future actions. It refers to the process of determining the position of a mobile robot with respect to its environment~\cite{huang1999robot}. One way for a robot to achieve localization is through odometry~\cite{borenstein1996sensors}, which involves estimating the position based on the rotation of the robot's wheels. 

The typical scenario for robot localization involves utilizing a map of the environment. Additionally, the robot has sensors that observe the environment and its own movement. Thus, the problem of localization involves estimating both the position and orientation of the robot within that map, using the information collected from these sensors. Not having an available map is another typical scenario. In such cases, Simultaneous Localization and Mapping (SLAM) \cite{durrant2006simultaneous,thrun2007simultaneous} can be applied. Consequently, while the robot navigates through the environment, it creates a map while simultaneously attempting to localize itself using that obtained map. 

Solving the localization problem and generating accurate maps become particularly complex when in indoor or outdoor environments. Indoor environments, such as an apartment, are typically flat and well-known environments from which a map can be easily obtained. In most cases, using a LiDAR sensor to produce the map and wheel odometry is sufficient for localization indoors \cite{tee2021lidar}. On the other hand, outdoor environments, such as a forest, are irregular environments. In these cases, it is necessary to employ other sensors such as depth cameras, 3D LiDAR, and GNSS. Therefore, the same sensors and algorithms cannot be used in all environments.

This paper describes the process undertaken to develop a mobile mapping system (MMS) for ROS 2-based robot localization based on various sensors for both indoor and outdoor scenarios. 
The MMS prototype proposed here is called VAULT, which stands for Visual and Autonomous Localization Kit. This prototype consists of cameras (depth and tracking), GNSS (Global Navigation Satellite System), and IMU (Inertial Measurement Unit). 
The information obtained from all the sensors can also be fused to obtain more accurate odometry. Various algorithms, such as Kalman filters \cite{sasiadek2002sensor}, are applied for this purpose. Lastly, with the depth camera information, Visual SLAM (VSLAM) \cite{fuentes2015visual} can be performed. VSLAM involves creating 2D and 3D maps based on depth information. With all these capabilities, a mobile robot can be localized in outdoor environments such as forests or crop fields.


\paragraph{Contributions}
The aim of VAULT project is to promote sustainability in agricultural, forestry, livestock, and mining activities through autonomous mobile robotic systems. The fundamental pillar of an autonomous system is the use of precise localization technologies that minimize errors in positioning tasks. Therefore, the main objective of this project is to create a functional prototype that offers precise localization for autonomous robots. This prototype will enable accurate localization for robotic platforms such as autonomous vehicles, agricultural robots, or rovers. To achieve this, it will gather localization information by processing data from various sensors, including cameras, GNSS, and IMU. Cameras will be used to generate visual odometry and Visual SLAM (VSLAM). The IMU (accelerometer and gyroscope) will provide corrections to improve the results. GNSS can be utilized to obtain more accurate odometry through the use of Real-Time Kinematics (RTK). In cases where satellite interference occurs, visual odometry can be used as a standalone alternative.

\section{Related works}

This section provides an overview of the recent developments and solutions to outdoor robot localization using complete devices composed of different hardware components. These devices are also known as mobile mapping systems (MMS) in the literature.

In \cite{shamseldin2018slam}, an MMS for an indoor robot is presented. This MMS is composed of a 3D LiDAR and a mini-computer that is in charge of computing SLAM. Another example of MMS for indoor robots is presented in \cite{yan2022real}, where a localization for agricultural robots is proposed. It is based on mapping utilizing multi-sensor fusion and visual odometry, IMU data and wheel odometry. The solution utilizes two cameras, an RGB-D camera and an Intel tracking camera; the wheel odometry of the Kobuki robot; and IMU data.

On the other hand, Elhashash discussed MMS as discussed in their comprehensive review \cite{elhashash2022review}. The MMS is defined as a system that combines LiDAR technology, a set of cameras, and sensors for positioning and georeferencing, such as the global navigation satellite system (GNSS), along with an inertial measurement unit (IMU).


Firstly, it is reviewed a commercial Leica Pegasus TRK100. It is a compact and highly adaptable system with a design of just 14 kg and dimensions [L/W/H] 70cm, 33cm and 49cm. It can be effortlessly installed and operated by a single person, streamlining the data collection process. The device boasts a modular camera system, allowing integration of up to four additional 24MP camera pairs. This configuration enables capturing front, side, and pavement angles for texture analysis, as well as intrinsic calibration for stitch-free panorama imagery. It also incorporates an Optical DMI, which eliminates the standard slip error of wheel-based DMIs and a dual scanner system. This device outperforms in hardware a regular MMS in characteristics and cost. 

Secondly, it reviews a commercial Hi-Scan-C device such as the one proposed by hi-target\footnote{\url{https://en.hi-target.com.cn/industries/3d-scanning-aerial-mapping/673/}}, which proposes an MMS based on a LiDAR-based method that incorporates a spherical camera, enabling it to produce geo-coded images along with a dense point cloud. This system is mounted on different vehicle types, capturing data as the vehicle traverses the designated area. Each point in the point cloud contains precise distance, location, and angular measurements. This extensive point cloud can be utilized for various applications, including 3D modeling, topographic landscape analysis, street view mapping, and numerous other Geographic Information System (GIS) applications. Everything is packed in a 15 kg device. 

Thirdly it is reviewed the MX9\footnote{\url{https://geospatial.trimble.com/products-and-solutions/trimble-mx9}}. It is an advanced mobile mapping system designed to be mounted on a vehicle's roof, the MX9 swiftly captures laser scans and images, including panoramic and multi-angle views, while on the move. It is also equipped with state-of-the-art Trimble GNSS and Inertial technology, the MX9 ensures very high point cloud density and exceptionally detailed imagery. The dimensions of the device are 0.62 m x 0.55 m x 0.62 m. However, the 37 kg of this unit makes the device slightly heavier compared with previous units.

The MS-96\footnote{\url{https://viametris.com/ms-96/}} is a versatile MMS designed for both indoor and outdoor scanning. It can be utilized in various configurations, such as backpacks, frontpacks, or car-mounted setups. The core of the system comprises a high-resolution 96Mpx panoramic camera, complemented by two LiDAR sensors capable of capturing an impressive 960000 points per second. Additionally, the MS-96 features an onboard Central Unit, a GNSS receiver, and a robust IMU with SLAM technology. It has a weight of 5.2 kg. 





These examples represent just a fraction of the diverse applications of mobile mapping systems across various industries and fields, showcasing their versatility and importance in modern mapping and application in outdoor jobs. Thus, this initial analysis enabled us to identify the strengths and limitations of current systems, understand the latest advancements in the field, and recognize potential gaps in the market.

\section{Materials and Methods}

The Materials and Methods section of this paper aims to detail the components and processes used to develop the VAULT prototype. The section is divided into the following subsections: Sensors in robot localization, Hardware, and Software.

\subsection{Sensors}

There is a large number of sensors that can be employed in localization systems. For instance, positioning and sensory data collection are two essential components used in a typical localization system in outdoor and indoor mobile robotics. Positioning sensors, that is GNSS, are used to obtain the geographical positions and motion of the sensors, that is IMU, are used to obtain relative inertial data. However, there are other sensors that can provide data about the environment such as cameras and LiDARs.

The GNSS receivers are one of the most used solutions for outdoor environments. They can be used to estimate absolute position, velocity and elevation in open areas within a global coordinate system. They receive signals from different navigational satellites and perform trilateration to calculate the real-time positions. These satellites mainly belong to the GPS developed by the United States, the GLONASS (Globalnaya Navigatsionnaya Sputnikovaya Sistema) developed by Russia, the Galileo built by the European Union, and the BeiDou system developed by China \cite{grewal2020global}. The GNSS usually presents certain errors in the calculations due to depending on external sources of data. However, this error can be improved by using high-precision GNSS receivers that use RTK (Real-Time Kinematic). The use of RTK can improve the positioning accuracy to decimeters and centimeters \cite{gan2007implement}.

IMU sensors consist of an accelerometer and a gyroscope which it uses to sense acceleration and angular
velocity. They record the relative position of the orientation and directional acceleration with respect to the initial position. The position and orientation data can be computed through dead reckoning approaches \cite{noureldin2012fundamentals}. Unlike GNSS, it does not depend on external sources. However, accumulated errors may produce drifts to their true positions.

Wheel encoders are classic sensors used in indoor environments. They are commonly used along with IMU. Odometry can be computed using the actual number of revolutions of the wheels obtained from the encoders \cite{ben2018robotic}. The main disadvantage is that it depends on the wheel size and is coupled in each robot.

LiDAR is an optical sensor that uses directional laser beams to measure the distances of objects. In indoor environments, 2D LiDAR is commonly used to perform SLAM~\cite{xuexi2019slam}. In this case, the LiDAR is situated in the base of the robot at a certain distance from the floor. However, 3D LiDARs are been used in outdoor environments \cite{zakaria2022ros,pfrunder2017real} instead of 2D LiDAR due to the irregularities of these scenarios.

Finally, cameras are the most popular sensor for data collecting. They are intended to acquire images at a high frame rate (30–60 frames per second) that can be used to perform visual odometry~\cite{nister2004visual} and Visual SLAM (VSLAM)~\cite{fuentes2015visual}. There are several cameras used: monocular, binocular, RGB-D and fisheye.

\subsection{VAULT Hardware}

\begin{figure}[!th]
  \begin{center}
\includegraphics[width=0.47\textwidth]{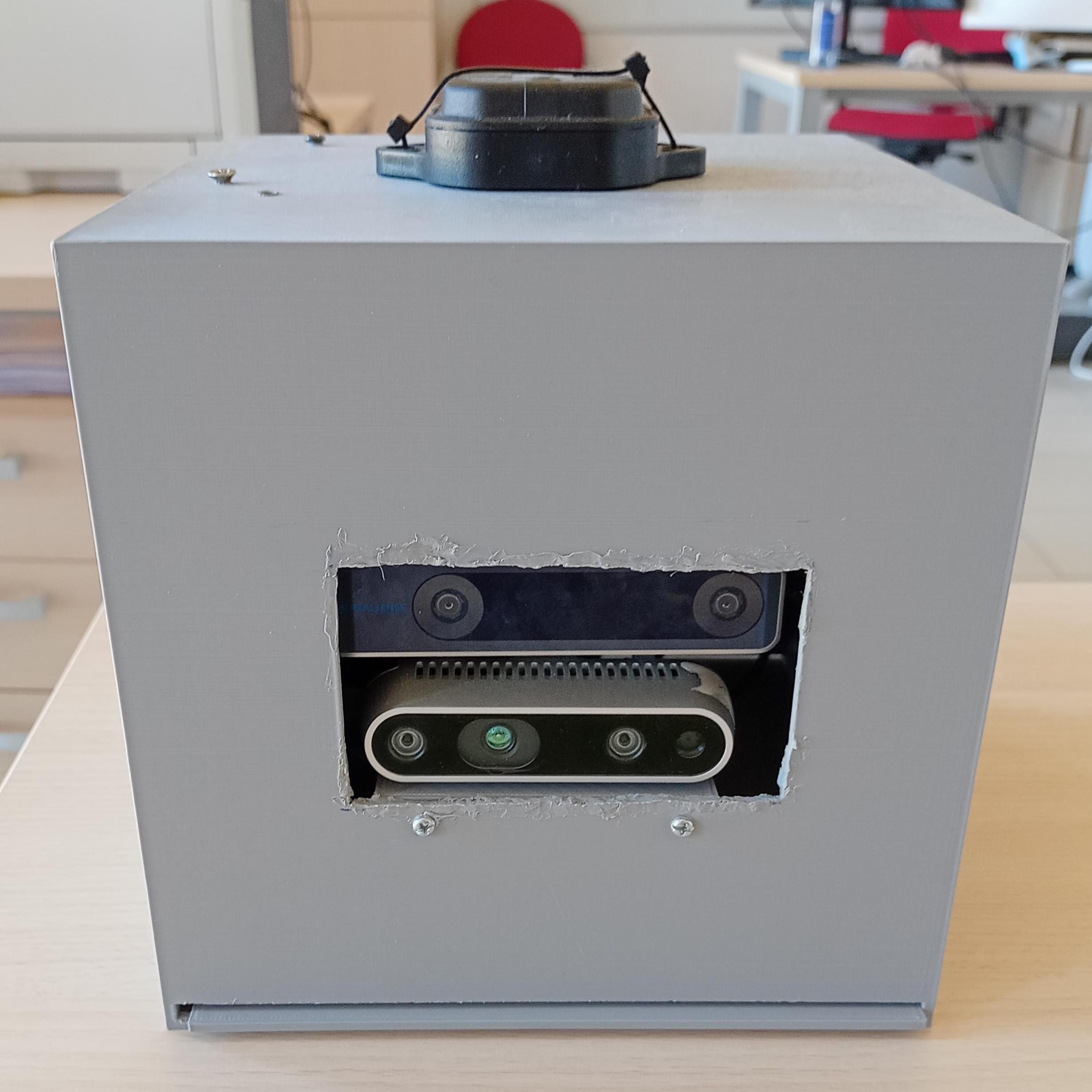}
\hspace{0.5cm}
\includegraphics[width=0.47\textwidth]{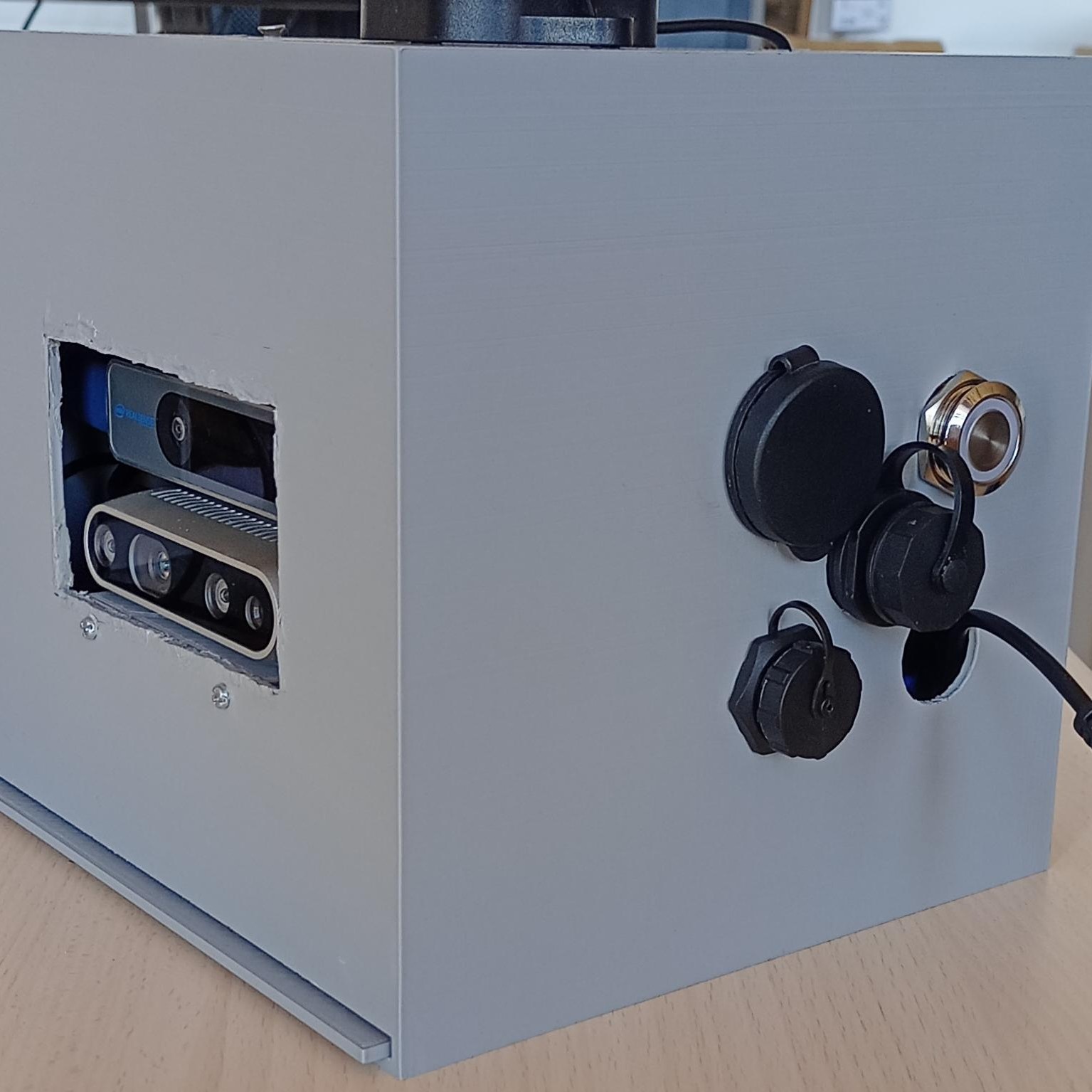}
\caption{Final version of the prototype VAULT.} \label{fig:VAULT_real}
  \end{center}
\end{figure}

VAULT prototype is presented in Figure~\ref{fig:VAULT_real}. The hardware components in this prototype, which are inside the 3D printed box presented in the figure, were carefully selected to ensure optimal performance and reliability in the autonomous robotic system. However, the authors follow a sensorization strategy to enhance the capabilities of the VAULT system. By incorporating redundancy sensors, the system should gather extra information about its surroundings and internal states, enabling a deeper information overview of the operating environment.

At the core of the hardware setup is a mini-computer platform BMAX B2Plus, equipped with an Intel Celeron processor running at 2.6 GHz, 8 GB of RAM, a 256 GB SSD, and the integrated Intel UHD Graphics 600. This computing platform serves as the brain of the autonomous system, enabling real-time decision-making and control. In order for the prototype to be portable and independently tested from a robot, a battery is deployed to power the hardware components. The chosen battery is a 12-volt, 60 Ah lithium battery.

The system includes two cameras: the Intel D435i camera and the Intel T256 camera. The Intel D435i camera is known for its versatility and accuracy in depth sensing. It combines an RGB camera with an infrared projector and infrared sensors, enabling it to capture both color and depth information. With a depth range of approximately 0.2 to 10 meters, the D435i is suitable for various applications, including robotics, 3D scanning, and augmented reality. It also comes with an integrated IMU (Inertial Measurement Unit) with 6 Degrees of Freedom.

On the other hand, the Intel T256 camera is specifically designed for high-precision depth perception in industrial environments. It utilizes Time-of-Flight (ToF) technology to capture depth information with exceptional accuracy and resolution. The T256 offers fisheye images and a longer depth range compared to the D435i, reaching up to 6 meters. It is well-suited for applications such as warehouse automation, object tracking, and obstacle detection. The T256 camera also features an integrated vision processor, offloading the processing workload from the host system.


To further enhance localization accuracy, the system integrates a Ublox C099-F9P GNSS. It is a GNSS receiver, coupled with Real-Time Kinematic (RTK) capability, which provides precise positioning information. This allows the system to generate more accurate odometry and compensate for any discrepancies that may arise from other sensors or environmental conditions.

Another crucial hardware component is the HFI-A9 IMU, which includes an accelerometer, a gyroscope and a magnetometer. The IMU captures information about the 9-axis: 3-axis gyroscope, 3-axis accelerometer, and 3-axis magnetometer. The gyroscope range of the device is ±2000°/s, while the acceleration range is ±8g. Additionally, the magnetometer range is ±1.3 Gauss. In terms of accuracy, the device offers a static angle accuracy of 0.1° and a dynamic angle accuracy of 0.5°. The return rate or baud rate is set at 300Hz/921600.

The complete VAULT prototype showcases a remarkably lightweight design with a total weight of just 2.1 kg. This is the result of the integration process of its hardware components, ensuring a compact and highly portable mobile mapping system. The lightweight nature of VAULT not only enhances its maneuverability and agility but also makes it a practical solution for deployment across a wide range of autonomous mobile robots, enabling integration without compromising the overall performance and capabilities of the host platform.

\subsection{VAULT Software}

VAULT utilizes ROS 2, the new version of ROS (Robot Operating System). This means that it can be used in any ROS 2-based robot, allowing it to produce robust localization data in outdoor and indoor environments. This software section is divided into the following subsections: the software used in localization and the simulation of VAULT. 

\subsubsection{Localization Software}

\begin{figure}[!th]
\centering
\includegraphics[width=1.0\textwidth]{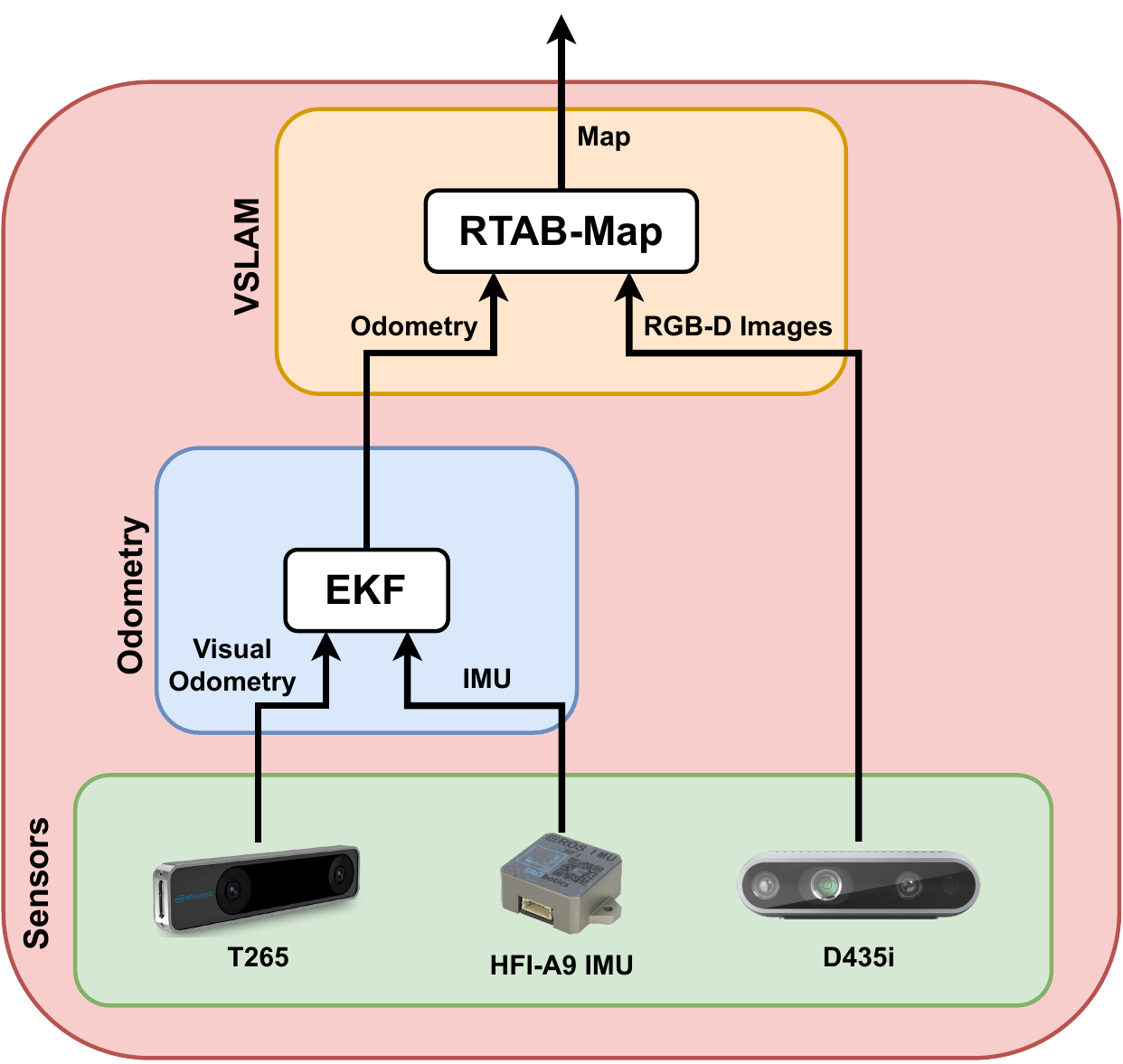}
\caption{VAULT's architecture, including sensors and software.} \label{fig:soft_arch}
\end{figure}

VAULT employs different software solutions to obtain highly accurate and reliable localization for its mobile mapping system (MMS). As a key feature of its design, VAULT strives for independence from the specific robot being utilized, and therefore, it does not rely on traditional wheel odometry. Instead, VAULT harnesses the power of cutting-edge technologies such as visual odometry, VSLAM (Visual Simultaneous Localization and Mapping), and sensor fusion to achieve precise localization results. The software architecture of VAULT, as depicted in Figure~\ref{fig:soft_arch}, showcases the integration of these essential components.

Visual odometry, a fundamental element of VAULT's localization strategy, is acquired through the use of the Intel T265 camera, which is well-known for its outstanding tracking capabilities. The Intel RealSense SDK\footnote{\url{https://github.com/IntelRealSense/librealsense/tree/master}} serves as the software interface for the camera, enabling the extraction of accurate visual odometry data. This visual odometry data is further enriched and refined through the integration of HFI-A9 IMU data. The Extended Kalman Filter (EKF)~\cite{ribeiro2004kalman}, a widely adopted estimation algorithm, is employed for this sensor fusion process. By incorporating information from both visual odometry and IMU data, VAULT is able to enhance the robustness and accuracy of its localization output. The ROS 2 package, robot\_localization\footnote{\url{http://docs.ros.org/en/noetic/api/robot\_localization/html/index.html}}, proves to be a valuable asset in this step, providing the implementation of the EKF and streamlining the fusion process.

The end result of the sensor fusion process is the generation of a highly reliable and precise odometry estimate for VAULT. This integrated odometry data forms the backbone of the mapping functionality of the system, making it an indispensable component for successful VSLAM.

The mapping capabilities of VAULT are carried out through the utilization of VSLAM, which is realized with the help of the RTAB-Map (Real-Time Appearance-Based Mapping) library~\cite{labbe2019rtab}. RTAB-Map enables VAULT to construct a detailed and accurate map of the environment while simultaneously localizing the robot within that map. For this purpose, RTAB-Map leverages not only the odometry data obtained from the sensor fusion but also RGB images and depth images captured by the Intel D435i camera. By fusing information from these diverse sources, VAULT gains a comprehensive understanding of its surroundings and builds a dynamic map that allows it to navigate efficiently and effectively.

Ublox C099-F9P GNSS data can be used in different ways. On one hand, it can be fed into the RTAB-Map to the VSLAM. On the other hand, it can be used to produce another odometry source that can be fed into the sensor fusion. However, the Ublox C099-F9P may produce noise data in indoor environments which means using only the inertial and visual data in the localization process of indoor environments.

In summary, VAULT plays a pivotal role in the success of the mobile mapping system. The integration of visual odometry, VSLAM, and sensor fusion techniques through sophisticated software libraries and algorithms in one MMS ensures that a ROS 2-based robot achieves exceptional localization accuracy and empowers it to create detailed and informative maps of its operating environment. This potent combination of hardware and software puts VAULT at the forefront of modern mapping and localization solutions, making it an invaluable tool for a wide range of applications, from autonomous navigation to environmental monitoring and exploration.

\begin{figure}[!th]
\centering
\includegraphics[width=0.83\textwidth]{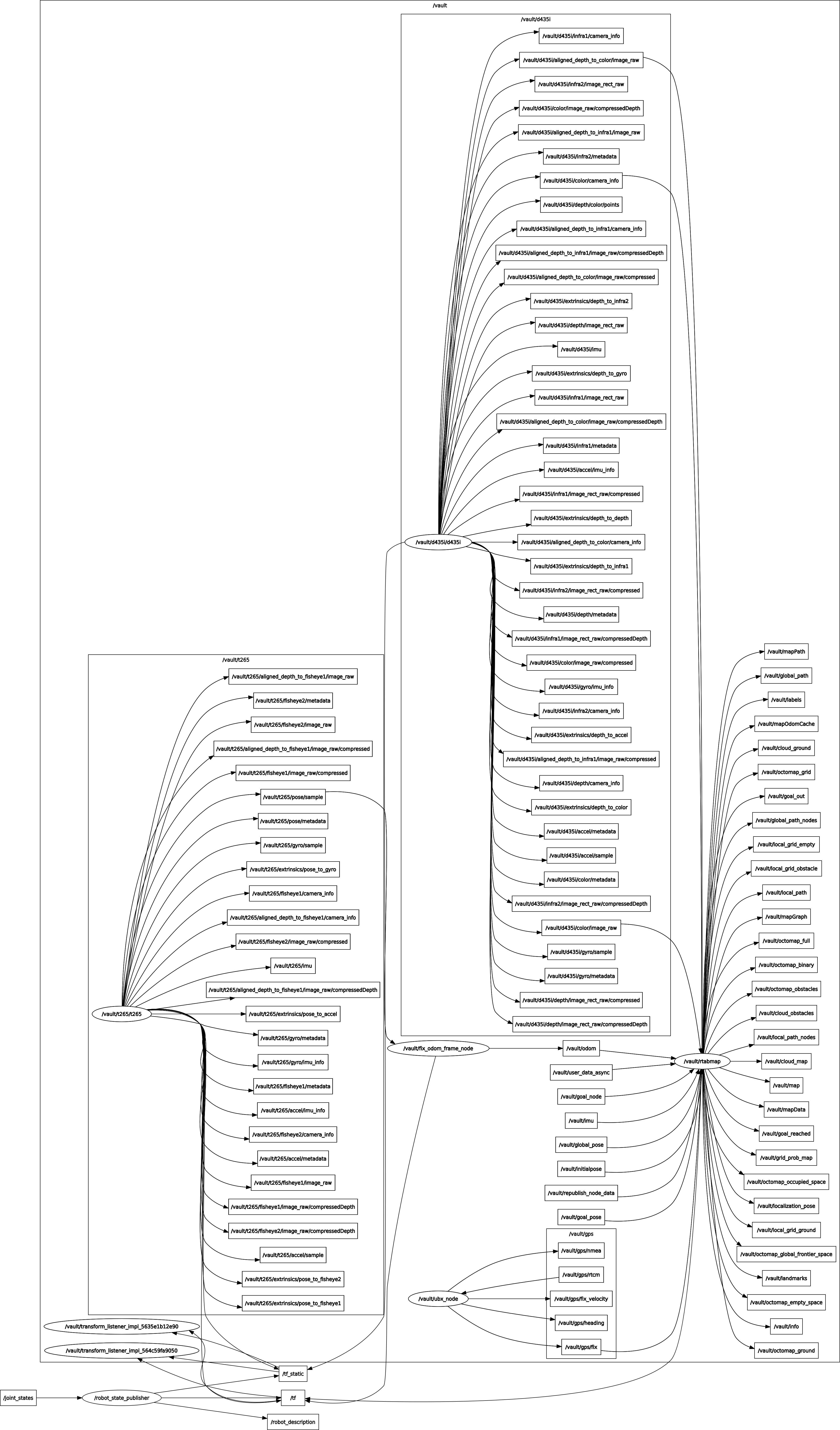}
\caption{Rosgraph of the prototype VAULT showing ROS 2 nodes and topics of the sensors and software tools.} \label{fig:rosgraph}
\end{figure}

The resulting $rosgraph$ (the graph that shows the peer-to-peer network of ROS processes that are processing data together) generated from the VAULT device is presented in Figure~\ref{fig:rosgraph}. This graph shows the ROS 2 nodes and topics of the Intel T265 camera, the Intel D435i camera, the HFI-A9 IMU, the robot\_localization EKF, the RTAB-Map VSLAM and the GNSS.

\subsubsection{Simulation}

\begin{figure}[!th]
\includegraphics[width=\textwidth]{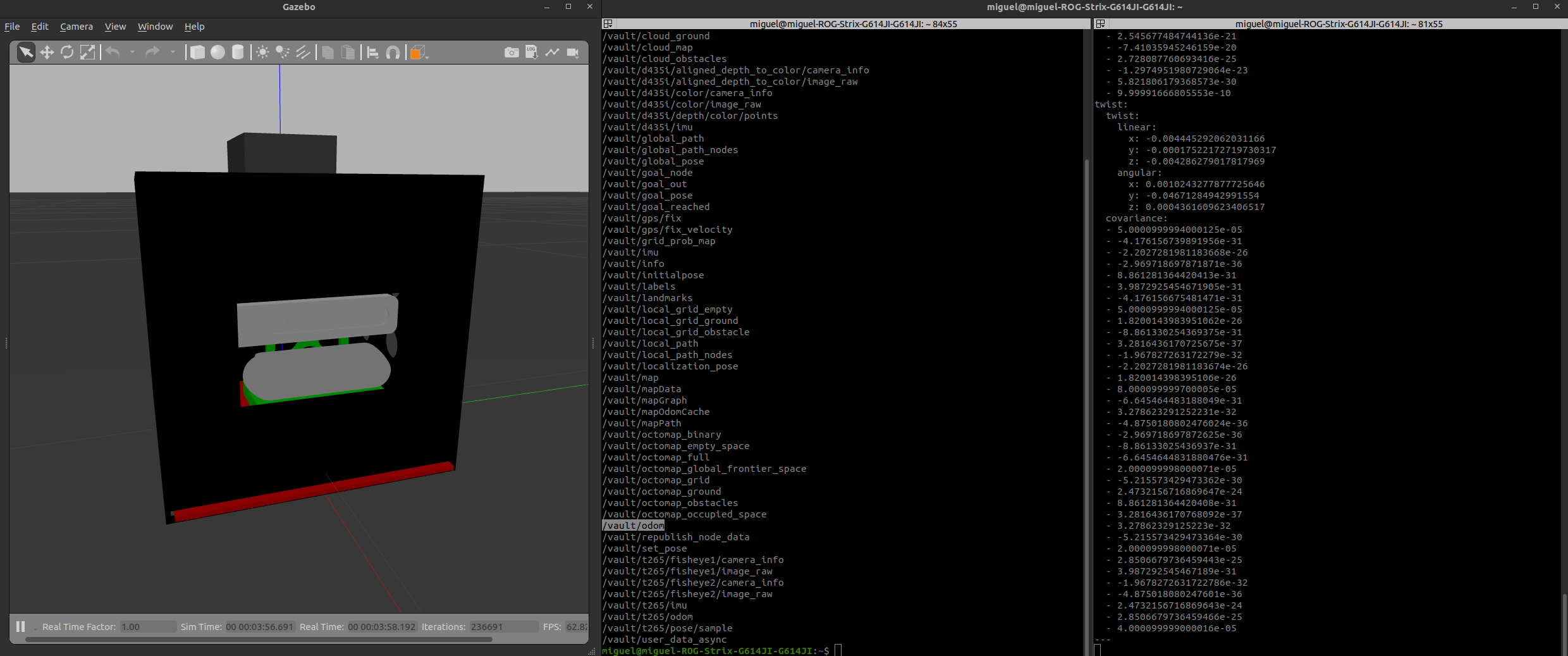}
\caption{Simulated version of VAULT prototype in Gazebo.} \label{fig:gazebo_VAULT}
\end{figure}

The simulated version of VAULT, implemented in Gazebo as shown in Figure~\ref{fig:gazebo_VAULT}, represents a significant step in the development and testing of the proposed MMS. By leveraging Gazebo's capabilities, the simulation faithfully replicates the behavior and functionality of VAULT in a virtual environment. This allows researchers and developers to thoroughly evaluate and fine-tune the system's performance without the constraints and risks associated with real-world testing.

One of the key aspects of the simulated version is the integration of virtual sensors, mirroring the ones present in the physical VAULT system. These simulated sensors include the Intel T265 camera and the Intel D435i camera, which are essential for visual odometry and VSLAM, respectively. The use of these simulated sensors enables the accurate emulation of data acquisition and processing, closely resembling the behavior of the physical sensors. This approach ensures that the algorithms and software developed for VAULT can be tested and validated in a controlled and reproducible manner within the Gazebo environment.

The simulated version of VAULT has been put through rigorous testing in various simulated outdoor environments. For this purpose, a simulated rover, described in \cite{ros2_rover_2021}, serves as a suitable platform for testing VAULT's capabilities. The rover in the simulation provides a versatile and representative platform that closely mimics the behavior and mobility of the actual robot used in conjunction with VAULT.

The use of Gazebo and the simulated rover enables researchers to evaluate VAULT's performance under diverse environmental conditions and challenging scenarios. This allows for systematic testing of different algorithms, tuning of parameters, and validation of localization accuracy and mapping fidelity. Additionally, the simulated setup offers the advantage of time-efficient testing, as simulations can be easily repeated and modified, leading to a more streamlined development and optimization process.

Furthermore, by using a simulated outdoor environment, researchers can push VAULT to its limits without the risk of damaging the physical system or encountering dangerous situations. This facilitates the identification of potential weaknesses or areas for improvement in the mapping and localization algorithms, thereby guiding the refinement of VAULT's software and hardware components.

In conclusion, the simulated version of VAULT in Gazebo, with its integration of virtual sensors and testing with the simulated rover, presents a powerful and safe approach for developing, evaluating, and refining the mobile mapping system. It serves as a vital step in the iterative process of creating a robust and reliable mapping solution, making VAULT better equipped to face real-world challenges and applications in various outdoor scenarios.

\section{Results and Discussion}

In the evaluation of VAULT's performance, three different real-world scenarios were carefully designed to provide a comprehensive assessment of the mobile mapping system's capabilities.


\begin{figure}[!bh]
    \centering
    \begin{subfigure}[t]{1.0\linewidth}
        \centering
        \includegraphics[width=1.0\linewidth]{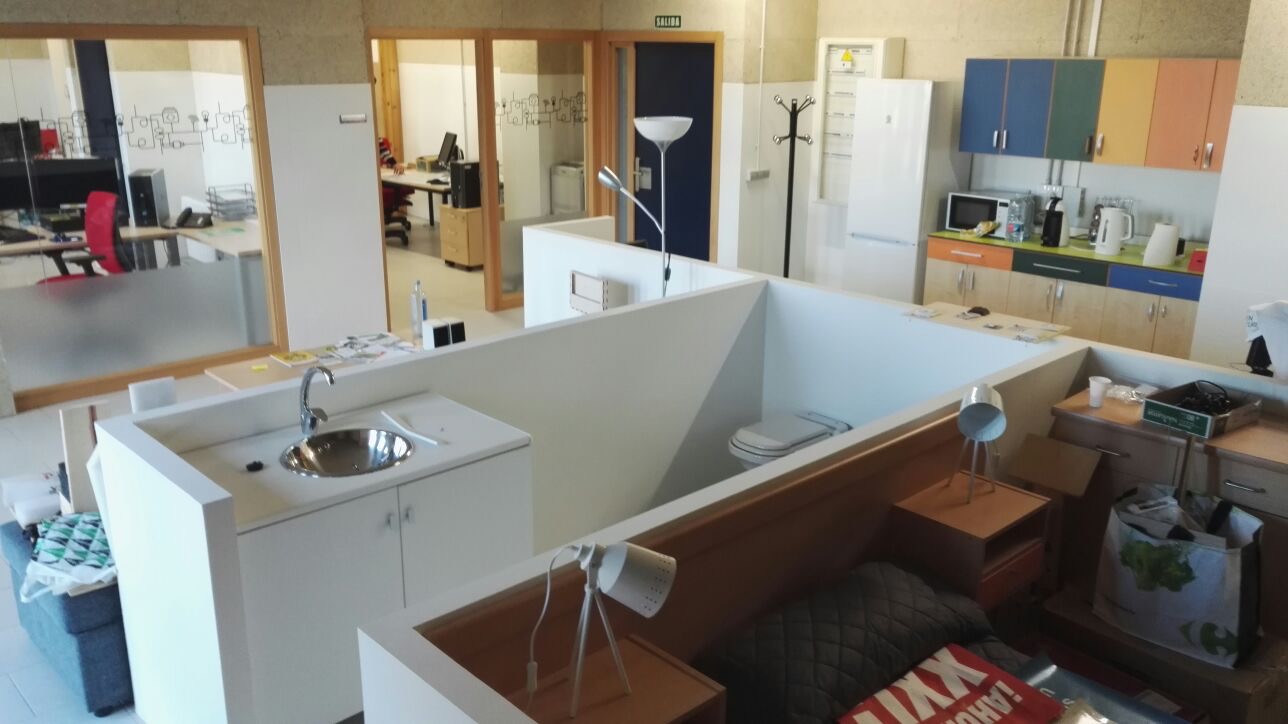}
        \caption{Leon@home Testbed apartment.}
    \end{subfigure}
    \begin{subfigure}[t]{.47\linewidth}
        \centering
        \includegraphics[width=1.0\linewidth]{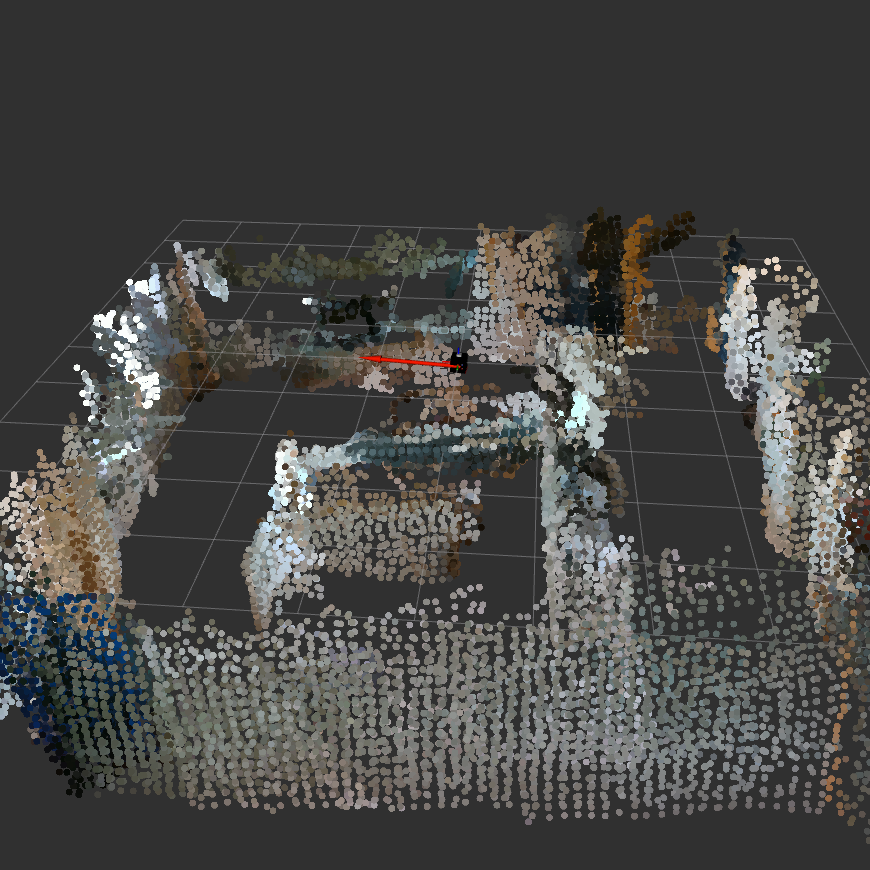}
        \caption{RViz visualization of VAULT in Leon@home Testbed apartment.}
    \end{subfigure}
    \hspace{0.5cm}
    \begin{subfigure}[t]{.47\linewidth}
        \centering
        \includegraphics[width=1.0\linewidth]{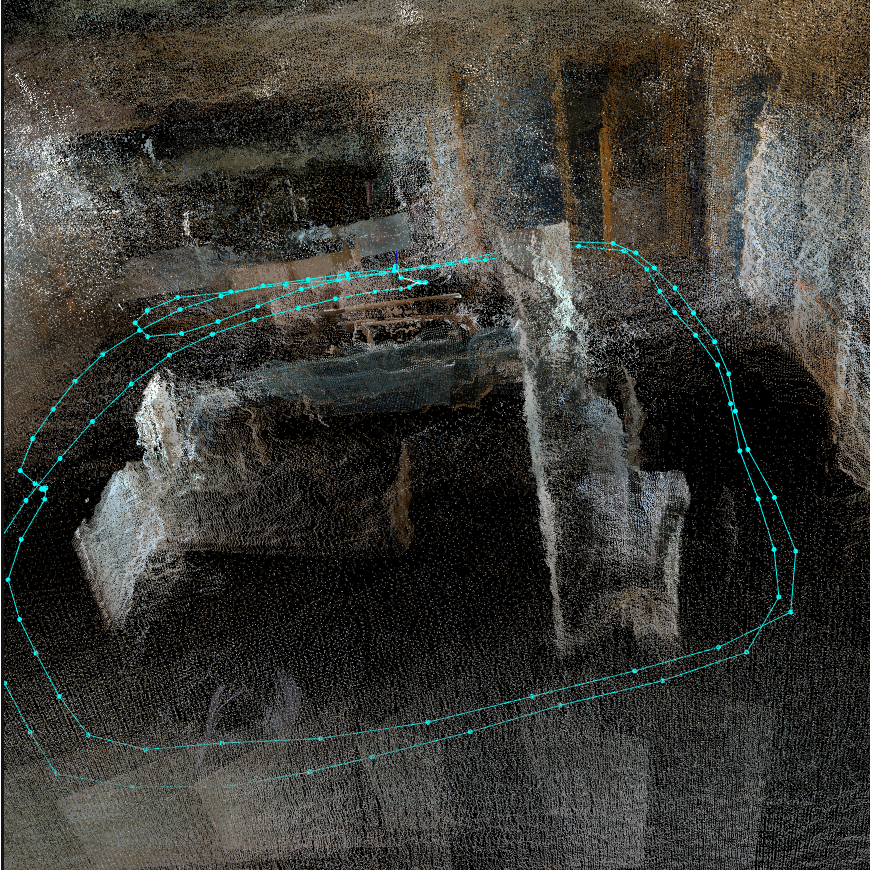}
        \caption{Rtabmapviz visualization of VAULT in Leon@home Testbed apartment.}
    \end{subfigure}     
    \caption{First scenario of experimentation in an indoor apartment, the Leon@home Testbed apartment.}\label{fig:experiment_scenarios_1}
\end{figure}

The first scenario, which is presented in Figure~\ref{fig:experiment_scenarios_1}, took place within the confines of the Leon@home Testbed, an indoor environment constructed as a single-bedroom mock-up home, occupying an 8m x 7m area. The testbed was thoughtfully partitioned into different functional spaces using 60cm high walls, creating distinct areas such as a Kitchen, living room, bathroom, and bedroom. This controlled indoor setting allowed researchers to assess VAULT's ability to map within complex indoor spaces, where the presence of walls and varied furniture arrangements pose challenges for autonomous mapping and localization systems.

The second scenario involved an indoor multi-floor environment, which presented additional complexities compared to the first scenario. This setup consisted of two interconnected corridors by two staircases. Localizing and mapping through corridors and dealing with staircases are particularly demanding tasks for mobile mapping systems. By subjecting VAULT to this scenario, researchers aimed to evaluate its adaptability and robustness in dealing with vertical movements and varying floor layouts.

The third scenario introduced an entirely different outdoor environment, specifically the surroundings of a building. The outdoor scenario offered unique challenges, including varying terrain, changing lighting conditions, and potential occlusions caused by natural and man-made objects. Mapping and localizing in such environments require sophisticated algorithms capable of handling complex 3D structures and accurately estimating positions even under dynamic lighting and weather conditions.

Conducting experiments in these three diverse scenarios allowed researchers to assess the versatility and generalizability of VAULT's mapping and localization capabilities. By analyzing its performance across different environments, VAULT's strengths and weaknesses could be identified, leading to targeted improvements in its software and hardware components. Additionally, the data gathered from these experiments offered valuable insights into the system's limitations and potential areas for future research, paving the way for the continued advancement of VAULT and its applications in real-world scenarios.

\subsection{Lessons Learned}

During the course of the experiments, several challenges and areas for improvement were identified in different aspects of VAULT's development and performance.

One of the primary challenges encountered during the hardware integration process was optimizing the space to accommodate all the necessary boards and wires. As VAULT aims to be a compact and efficient mobile mapping system, achieving an effective hardware layout was crucial. The 3D printing process played a significant role in fabricating custom components to fit the specific requirements of VAULT's hardware configuration.

Another critical aspect that came to light during the experiments was the need for an energy consumption study. Understanding the power requirements of VAULT and its individual sensors, including the Intel D435i camera, the Intel T265 camera, the HFI-A9 IMU, and the Ublox GNSS, is essential for ensuring adequate battery management. Being independent of the robotics platform, VAULT needs to control the battery output precisely to meet the power demands of its sensors and maximize operational efficiency.

The sensorization strategy employed in VAULT proved to be a valuable feature, enabling the system to gather data from multiple sources, such as fisheye images, RGB images, depth images, GNSS data, and IMU data. This wealth of sensory information enhances VAULT's localization capabilities and paves the way for more sophisticated localization strategies by utilizing diverse data sources and producing richer odometry information.

Finally, simulation played a vital role in refining the software sections of VAULT. By simulating all the sensors, including the HFI-A9 IMU, the Ublox GNSS, and the Intel D435i camera, researchers were able to efficiently develop and test different algorithms and software components. However, the simulation of the visual odometry obtained directly from the Intel T265 camera using the Gazebo P3D plugin presented some challenges that need to be addressed to ensure accurate and reliable simulation results.

\section{Conclusions}

This paper presents VAULT, a ROS 2-based Mobile Mapping System (MMS) designed to achieve precise and reliable localization for autonomous mobile robots. Throughout the development of VAULT, a series of experiments were conducted in three diverse scenarios, encompassing indoor and outdoor environments, multi-floor structures, and complex surroundings. These experiments provided invaluable insights into the system's capabilities, strengths, and areas for improvement.

One of the key highlights of VAULT is its independence from the specific robot platform, achieved through the integration of visual odometry, VSLAM, and sensor fusion techniques. By leveraging the power of the Intel T265 camera for visual odometry, the Intel D435i camera for VSLAM, and the HFI-A9 IMU, VAULT effectively generates highly accurate odometry data. The integration of these data streams through the Extended Kalman Filter (EKF) algorithm further enhances the system's localization accuracy and robustness.




While VAULT demonstrates impressive localization capabilities through the integration of visual odometry, VSLAM, and sensor fusion techniques, future works could explore the inclusion of more sensory data to further enhance its performance. For instance, incorporating more IMU data or data from new sensors like LiDAR, radar and thermal cameras. These new sensors could provide complementary information, especially in challenging environments with low visibility or dynamic obstacles. Besides more odometry sources, such as RGB-D odometry and stereo odometry, can be included in the sensor fusion process.

In addition to exploring the integration of more sensory data, the utilization of deep learning techniques presents an exciting avenue for future improvements in VAULT. Deep learning has shown remarkable success in various robotics tasks, such as object detection, semantic segmentation, and path planning. Integrating deep learning models into VAULT could enhance its ability to adapt to complex and dynamic surroundings. Thus, a multi-modal deep learning solution that uses the sensory data of VAULT could result in an improved version.

\section*{Funding}
Grant TED2021-132356B-I00 funded by MCIN/AEI/10.13039/501100011033 and by the ``European Union NextGenerationEU/PRTR''.

\section*{Acknowledgements}
Miguel \'{A}. Gonz\'{a}lez-Santamarta acknowledges an FPU fellowship provided by the Spanish Ministry of Universities (FPU21/01438).


%
%
%
\bibliographystyle{splncs04}
\bibliography{references}

\begin{thebibliography}{10}
\providecommand{\url}[1]{\texttt{#1}}
\providecommand{\urlprefix}{URL }
\providecommand{\doi}[1]{https://doi.org/#1}

\bibitem{ben2018robotic}
Ben-Ari, M., Mondada, F., Ben-Ari, M., Mondada, F.: Robotic motion and
  odometry. Elements of Robotics pp. 63--93 (2018)

\bibitem{borenstein1996sensors}
Borenstein, J., Everett, H., Feng, L., et~al.: Where am i? sensors and methods
  for mobile robot positioning. University of Michigan  \textbf{119}(120), ~27
  (1996)

\bibitem{durrant2006simultaneous}
Durrant-Whyte, H., Bailey, T.: Simultaneous localization and mapping: part i.
  IEEE robotics \& automation magazine  \textbf{13}(2),  99--110 (2006)

\bibitem{elhashash2022review}
Elhashash, M., Albanwan, H., Qin, R.: A review of mobile mapping systems: From
  sensors to applications. Sensors  \textbf{22}(11), ~4262 (2022)

\bibitem{fuentes2015visual}
Fuentes-Pacheco, J., Ruiz-Ascencio, J., Rend{\'o}n-Mancha, J.M.: Visual
  simultaneous localization and mapping: a survey. Artificial intelligence
  review  \textbf{43}(1),  55--81 (2015)

\bibitem{gan2007implement}
Gan-Mor, S., Clark, R.L., Upchurch, B.L.: Implement lateral position accuracy
  under rtk-gps tractor guidance. Computers and Electronics in Agriculture
  \textbf{59}(1-2),  31--38 (2007)

\bibitem{ros2_rover_2021}
Gonz{\'a}lez-Santamarta, M.{\'A}.: {ros2\_rover} (Jul 2021),
  \url{https://github.com/mgonzs13/ros2\_rover}

\bibitem{grewal2020global}
Grewal, M.S., Andrews, A.P., Bartone, C.G.: Global navigation satellite
  systems, inertial navigation, and integration. John Wiley \& Sons (2020)

\bibitem{huang1999robot}
Huang, S., Dissanayake, G.: Robot localization: An introduction. Wiley
  Encyclopedia of Electrical and Electronics Engineering pp. 1--10 (1999)

\bibitem{labbe2019rtab}
Labb{\'e}, M., Michaud, F.: Rtab-map as an open-source lidar and visual
  simultaneous localization and mapping library for large-scale and long-term
  online operation. Journal of Field Robotics  \textbf{36}(2),  416--446 (2019)

\bibitem{nister2004visual}
Nist{\'e}r, D., Naroditsky, O., Bergen, J.: Visual odometry. In: Proceedings of
  the 2004 IEEE Computer Society Conference on Computer Vision and Pattern
  Recognition, 2004. CVPR 2004. vol.~1, pp.~I--I. Ieee (2004)

\bibitem{noureldin2012fundamentals}
Noureldin, A., Karamat, T.B., Georgy, J.: Fundamentals of inertial navigation,
  satellite-based positioning and their integration. Springer Science \&
  Business Media (2012)

\bibitem{pfrunder2017real}
Pfrunder, A., Borges, P.V., Romero, A.R., Catt, G., Elfes, A.: Real-time
  autonomous ground vehicle navigation in heterogeneous environments using a 3d
  lidar. In: 2017 IEEE/RSJ International Conference on Intelligent Robots and
  Systems (IROS). pp. 2601--2608. IEEE (2017)

\bibitem{ribeiro2004kalman}
Ribeiro, M.I.: Kalman and extended kalman filters: Concept, derivation and
  properties. Institute for Systems and Robotics  \textbf{43}(46),  3736--3741
  (2004)

\bibitem{sasiadek2002sensor}
Sasiadek, J.Z.: Sensor fusion. Annual Reviews in Control  \textbf{26}(2),
  203--228 (2002)

\bibitem{shamseldin2018slam}
Shamseldin, T., Manerikar, A., Elbahnasawy, M., Habib, A.: Slam-based
  pseudo-gnss/ins localization system for indoor lidar mobile mapping systems.
  In: 2018 IEEE/ION Position, Location and Navigation Symposium (PLANS). pp.
  197--208. IEEE (2018)

\bibitem{tee2021lidar}
Tee, Y.K., Han, Y.C.: Lidar-based 2d slam for mobile robot in an indoor
  environment: A review. In: 2021 International Conference on Green Energy,
  Computing and Sustainable Technology (GECOST). pp.~1--7. IEEE (2021)

\bibitem{thrun2007simultaneous}
Thrun, S.: Simultaneous localization and mapping. In: Robotics and cognitive
  approaches to spatial mapping, pp. 13--41. Springer (2007)

\bibitem{xuexi2019slam}
Xuexi, Z., Guokun, L., Genping, F., Dongliang, X., Shiliu, L.: Slam algorithm
  analysis of mobile robot based on lidar. In: 2019 Chinese Control Conference
  (CCC). pp. 4739--4745. IEEE (2019)

\bibitem{yan2022real}
Yan, Y., Zhang, B., Zhou, J., Zhang, Y., Liu, X.: Real-time localization and
  mapping utilizing multi-sensor fusion and visual--imu--wheel odometry for
  agricultural robots in unstructured, dynamic and gps-denied greenhouse
  environments. Agronomy  \textbf{12}(8), ~1740 (2022)

\bibitem{zakaria2022ros}
Zakaria, W.N.W., Mahmood, I.A.T., Shamsudin, A.U., Rahman, M.A.A., Tomari,
  M.R.M.: Ros-based slam and path planning for autonomous unmanned surface
  vehicle navigation system. In: 2022 IEEE 5th International Symposium in
  Robotics and Manufacturing Automation (ROMA). pp.~1--6. IEEE (2022)

\end{thebibliography}

\end{document}